\documentclass[sigconf]{acmart}

\renewcommand\footnotetextcopyrightpermission[1]{} 

\usepackage{booktabs} 
\usepackage{graphicx}
\usepackage{subcaption}

\setcopyright{rightsretained}

\acmDOI{10.475/123_4}

\acmISBN{123-4567-24-567/08/06}

\acmArticle{4}
\acmPrice{5.00}

\begin{document}
\title{Plan Explanations as Model Reconciliation -- An Empirical Study}
\author{Tathagata Chakraborti, Sarath Sreedharan, Sachin Grover, Subbarao Kambhampati}  
\authornote{The first two authors contributed equally.}
\affiliation{
School of Computing, Informatics, and Decision Systems Engineering\\
Arizona State University, Tempe, AZ 85281 USA\\[1ex]
}
\email{ tchakra2, ssreedh3, sgrover6, rao  @ asu.edu}




\newcommand{\note}{\textcolor{red}}
\newcommand{\plc}{\textcolor{red}{[placeholder]}}

\begin{abstract}
Recent work in explanation generation for decision making agents has looked at how unexplained behavior of autonomous systems can be understood in terms of differences in the model of the system and the human's understanding of the same, and how the explanation process as a result of this mismatch can be then seen as a process of reconciliation of these models. 
Existing algorithms in such settings, while having been built on contrastive, selective and social properties of explanations as studied extensively in the psychology literature, have not, to the best of our knowledge, been evaluated in settings with actual humans in the loop. 
As such, the applicability of such explanations to human-AI and human-robot interactions remains suspect. 
In this paper, we set out to evaluate these explanation generation algorithms in a series of studies in a mock search and rescue scenario with an internal semi-autonomous robot and an external human commander.
We demonstrate to what extent the properties of these algorithms hold as they are evaluated by humans, and how the dynamics of trust between the human and the robot evolve during the process of these interactions.
\end{abstract}

%
%



\maketitle

\section{Introduction}

The issue of explanations for AI systems operating alongside or with humans in the loop has been a topic of considerable interest of late \cite{darpa,xai}, especially as more and more AI-enabled components get deployed into hitherto human-only workflows. 
The ability to generate explanations holds the key \cite{slate,langley2017explainable} towards acceptance of AI-based systems in collaborations with humans. 
Indeed, in many cases, this may even be {\em required} by law \cite{2016arXiv160608813G}. 

\begin{figure}
\includegraphics[width=\columnwidth]{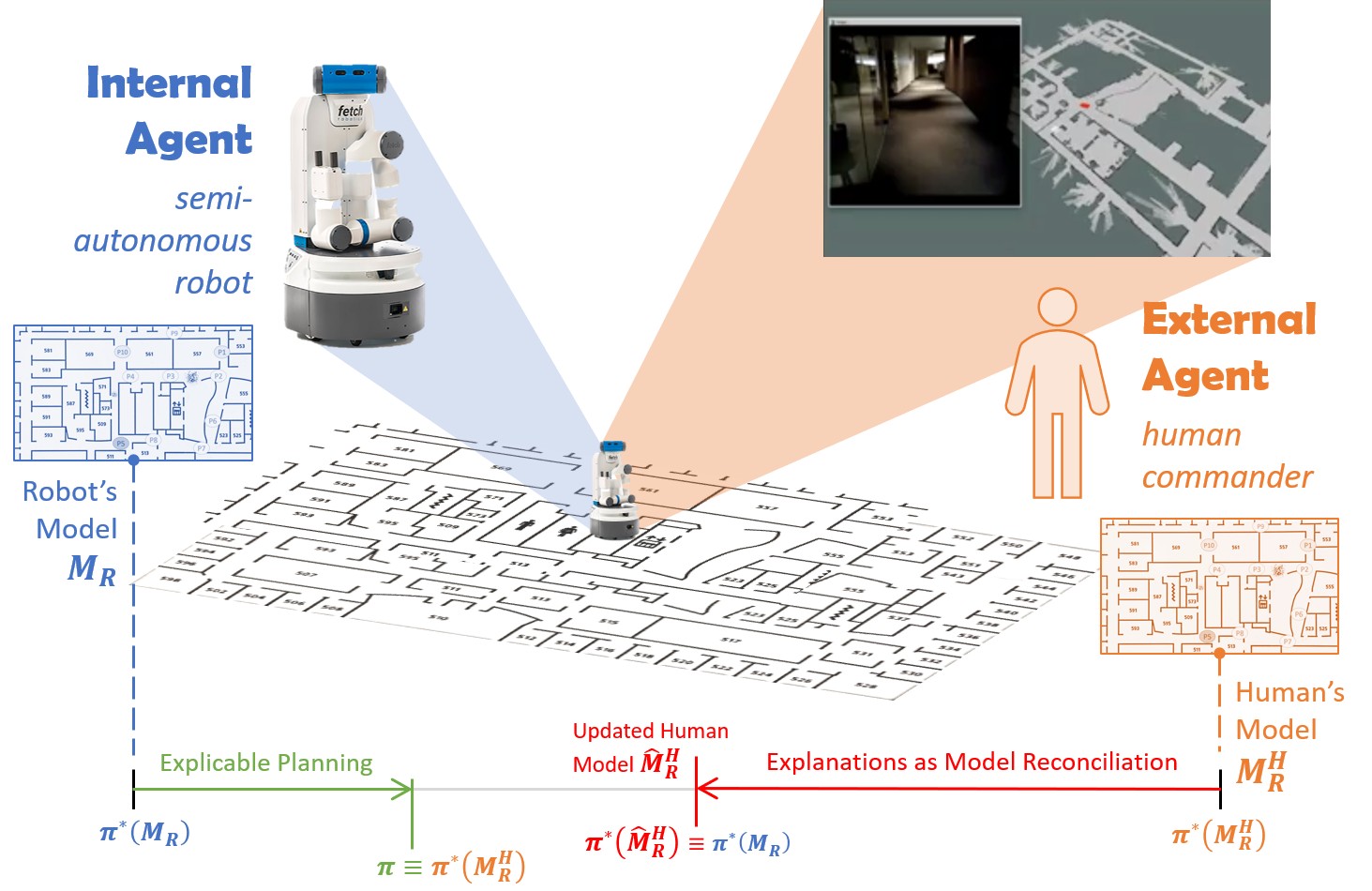}
\caption{A typical \cite{nancy} urban search and rescue (USAR) scenario with an internal semi-autonomous robot and an external human supervisor. Models (e.g. map of the environment) may diverge in the course of the operation due to the disaster. 
The robot can either choose to generate {\em explicable plans} by conforming to the expectations of the human in the loop or {\em explain} its plans to the human in terms of their model differences via a process called {\em model reconciliation}.}
\label{mrp}
\end{figure}

Of course, the answer to what constitutes a valid, or even useful, explanation largely depends on the type of AI-algorithm in question. 
Recent works \cite{explain,exp-fss,exp-fss-multi} have attempted to address that question in the context of human-robot interactions \cite{2017arXiv170704775C} by formulating the process of explaining the decisions of an autonomous agent as a {\em model reconciliation process} whereby the agent tries to bring the human in the loop to a shared understanding of the current situation so as to explain its decisions in that updated model. 
This is illustrated in Figure~\ref{mrp}. 
While these techniques have been developed on theories in the psychology literature \cite{Lombrozo2006464,lombrozo2012explanation} built on extensive studies in how humans explain behavior, none of these algorithms have, to the best of our knowledge, been evaluated yet with humans in the loop.
As such, it remains unclear whether the theoretical guarantees provided by explanations generated by such algorithms do, in fact, bear out during interactions with humans. 

The aim of this paper is then to provide an empirical study of the ``explanation as model reconciliation'' process, especially as it relates to a human-robot dyad in a mock up version of a typical search and rescue scenario (Section~\ref{usar}) which the authors in \cite{exp-fss,exp-fss-multi} have repeatedly used as an illustrative scenario.
But before we go there, we will provide a brief overview of explanations (Section~\ref{bho}) in the planning community and a glossary of terms (Section~\ref{glossary}) used throughout the rest of the discussion.

\section{\texorpdfstring{A Brief History of\\Explainable Planning}{A Brief History of Explainable Planning}}
\label{bho}

From the perspective of planning and decision making, the notion of explanations of the deliberative process of an AI-based system was first explored extensively in the context of {\em expert systems} \cite{moore1988explanation}.
Similar techniques have been looked at for explanations in case based planning systems \cite{kambhampati1990classification,Sørmo2005} and in interactive planning \cite{dave} where the planner is mostly concerned with establishing the correctness \cite{howey2004val} and quality \cite{sohBiaMcIAAAI2011,danmaga} of a given plan {\em with respect to its own model}. 
These explanation generation techniques served more as a debugging system for an expert user rather than explanations for situations generally encountered in everyday interactions, which may be referred to as {\em ``everyday explanations''} \cite{miller}. 
A key difference here is that the former is mostly algorithm dependent and explains the {\em how} of the decision making process whereas the latter, in addition, can model-based and hence algorithm independent and thus, in a sense, explain the {\em why} of a particular decision in terms of the knowledge that engendered it.
 
In \cite{asylum} authors argued that, in a classic case of ``inmates running the asylum'', most of the existing literature on explanation generation techniques for AI systems are based on the developer's intuitions rather than any principled understanding of the normative explanation process in interactions among humans as has been studied extensively in the fields of philosophy, cognitive psychology/science, and social psychology.
The authors note that the latter can be a valuable resource for the design of explanation generation methodologies in AI-based systems as well.

The authors in \cite{miller} state the three most important properties of explanations (as accounted for in the existing literature in the social sciences) as being (1) contrastive (so as to be able to compare the fact being questioned with other alternatives or foils); (2) selective (of what information among many to attribute causality of an event to); and (3) social (implying that the explainer must be able to leverage the mental model of the explainee while engaging in the explanation process).
In recent work on explanation generation for planners, authors in \cite{explain} expressed similar sentiments by arguing that the explanation process towards end users ``cannot be a soliloquy'' but rather a process of ``model reconciliation'' during which the system tries to bring the mental model ({\em social property}) of the user on the same page with respect to the plan being explained. 
The authors in \cite{explain} addressed the {\em contrastive property} by ensuring optimality of the plan being explained in the updated human mental model, and the {\em selectivity property} by computing the minimum number of updates required to realize the above constraint. 

In a related thread of work, researchers have looked recently at the idea of ``explicable'' planning \cite{exp-yu,zhang2016plan,exp-anagha} which aims to circumvent the need for explanations by instead having the robot sacrifice optimality in its own model and produce plans that are as close to optimal as possible in the mental model of the human in the loop. Of course, such plans may be too costly, or even infeasible, from the robot's perspective and as such the process of explicability and explanations form a delicate balancing act \cite{exp-fss} during the deliberative process of a decision making agent and forms a basis of an augmentative theory \cite{argue} of planning for an automated agent.

The process of explanations and explicability for task plans, in general, is also a harder process than in motion planning (c.f. recent works on ``legibility'' \cite{Dragan–2013–7671} and ``verbalization'' \cite{perera2016dynamic}) where acceptable behavior can be understood in terms of simple rules (e.g. minimizing distance to shortest path). 
In the case of task planning, human mental models are harder to acquire and thus must be {\em learned} \cite{exp-yu}.
Further, given a mental model of the user, it is still a challenge on how to leverage that model in the explanation process, keeping in mind the cognitive abilities and implicit processes and preferences of the human in the loop that are often very hard, or even impossible, to codify precisely in the task model itself.
Evaluation of learned mental models is out of scope of the current discussion, though readers are encouraged to refer to \cite{exp-yu} for related studies.
In this paper, we will focus only on known models, and explore how humans respond to these techniques in situations where these models diverge.
In the next section, we will describe some of the terms as it relates to explicable planning and plan explanations that will be used throughout the rest of the paper.

\begin{figure*}
\includegraphics[width=\textwidth]{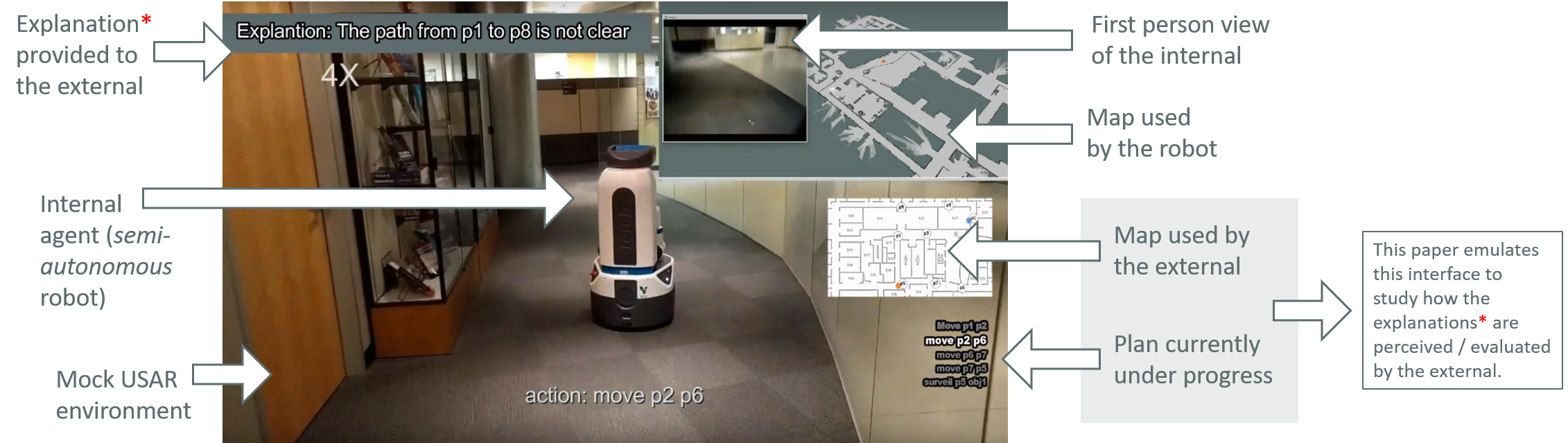}
\caption{Illustration of a simulated USAR setting for studying the human-robot relationship in a typical disaster response team. The external (human) supervisor has restricted access to a changing environment and may thus require explanations for plans that the (internal semi-autonomous) robot comes up with for tasks assigned to it. We make use of this setting to study the properties of ``model-reconciliation" explanations in a mock interface (Figure~\ref{interface2}) to the robot.}
\label{bhushi}
\end{figure*}

\section{Glossary of Terms}
\label{glossary}

Existing teamwork literature \cite{cooke2013interactive} on human-human and human-animal teams has identified characteristics of effective teams --
in terms of shared mental models \cite{Bowers,Mathieu2000} that contribute to team situational awareness \cite{nancy-new} and interaction \cite{nancy2013}.
Thus, it has been argued \cite{2017arXiv170704775C} that the ability to leverage these shared mental models, and reasoning over multiple models at a time, during the decision making process is critical to the effective design of cognitive robotic agents for teaming with humans.
The multi-model setting is illustrated in Figure~\ref{mrp} in the context of a search and rescue scenario (more on this later in Section~\ref{usar}) where the map of the environment shared across the robot and its operator diverge in course of operations.  
When making plans in such scenarios, the robot can choose to either (1) conform to human expectations, potentially sacrificing optimality in the process; or (2) preserve optimality and explain its plan (which may thus be inexplicable) in terms of the model differences (that causes this inexplicability). 
As explained before, the former process is described as explicable planning, while the latter is referred to as explanations as model reconciliation.

\subsection{Explicable Plans}

Let the model (which includes beliefs or state information and desires or goals as well as the action model) that the robot is using to plan be given by $\mathcal{M}_R$ and the human's understanding of the same be given by $\mathcal{M}^H_R$. 
Further, let $\pi^*(\mathcal{M}_R)$ and $\pi^*(\mathcal{M}^H_R)$ be the optimal plans in the respective models, and $C_{\mathcal{M}}(\cdot)$ be the (cost) function denoting the goodness of a plan in a model $\mathcal{M}$ (less the better).
When $\mathcal{M}^H_R \not= \mathcal{M}_R$, it is conceivable that $C_{\mathcal{M}^H_R}(\pi^*(\mathcal{M}_R)) > C_{\mathcal{M}^H_R}(\pi^*(\mathcal{M}^H_R))$ which constitutes an inexplicable behavior from the perspective of the human in the loop. 

\vspace{5pt}
\noindent {\bf In explicable planning}, the robot instead produces a plan $\pi$ such that $C_{\mathcal{M}^H_R}(\pi) \approx C_{\mathcal{M}^H_R}(\pi^*(\mathcal{M}^H_R))$, i.e. an explicable plan is equivalent (or as close as possible) to the human's expectation.

\subsection{Plan Explanations as Model Reconciliation}
\label{gloss}

The robot can, instead chose to stay optimal in its own model, and explain away the reasons, i.e. model differences, that causes its plan to be suboptimal in the human's mental model. 

\vspace{5pt}
\noindent {\bf The Model Reconciliation Problem (MRP)} involves the robot providing an explanation or model update $\mathcal{E}$ to the human so that in the new updated human mental model $\widehat{\mathcal{M}}^H_R$ the original plan is optimal (and hence explicable), i.e. $C_{\widehat{\mathcal{M}}^H_R}(\pi^*(\mathcal{M}^H_R)) = C_{\widehat{\mathcal{M}}^H_R}(\pi^*(\mathcal{M}_R))$\footnote{We refer to this constraint as the ``optimality condition'' in later discussions and the explanations that satisfy this condition is called complete explanations.}.

\vspace{5pt}
\noindent Of course, there may be many different types of these explanations, as explained below (terms reused from \cite{explain}. 

\subsubsection{Model Patch Explanations (MPE)}

Providing the entire model differences as a model update is a trivial solution. It satisfies the optimality criterion but may be too large from the point of communication when the robot has to operate with minimum bandwidth as well as cause loss of situational awareness and increased cognitive load on the part of the human by providing too much information that is not relevant to the plan being explained.

\subsubsection{Plan Patch Explanations (PPE)}

These restrict model changes to only those actions that appear in the plan. This kind of explanations do not satisfy the optimality criterion but ensure the executability of the given plan instead. Further, it may still contain information that is not relevant to explaining the original robot plan as opposed to the human expectation or foil.

\subsubsection{Minimally Complete Explanations (MCE)}

These explanations, on top of satisfying the optimality condition, also enforce $\min\mathcal{E}$.
This means MCEs not only make sure that the plan being explained is optimal in the updated model but also it is the minimum set of updates required to make this happen.
This is especially useful in reducing irrelevant information during the explanation process both from the perspective of the human as well as the robot when communication is expensive.

\subsubsection{Minimally Monotonic Explanations (MME)} 

Interestingly, MCEs can become invalid when combined, i.e. when multiple plans are being explained, the current MCE can make a previous one violate the optimality constraint. This leads to the notion of MMEs which guarantee that an explanation is always valid regardless of other plans being explained in the future (while at the same time revealing as little information as possible). 
This is especially useful in long term interactions in the human-robot dyad and is out of scope of the current study.

\subsection{Balancing Explicability and Explanations}
\label{balance}

Finally, as mentioned before, these ideas can come together whereby an agent can choose to trade off the cost of explanations versus the cost of producing explicable plans by performing model space search during the plan generation process \cite{exp-fss}.
In the following studies we simulate such an agent and generate plans that are either explicable, or optimal in the robot's model or somewhere in between (with an associated MCE, MPE or PPE).

\begin{figure*}
\includegraphics[width=\textwidth]{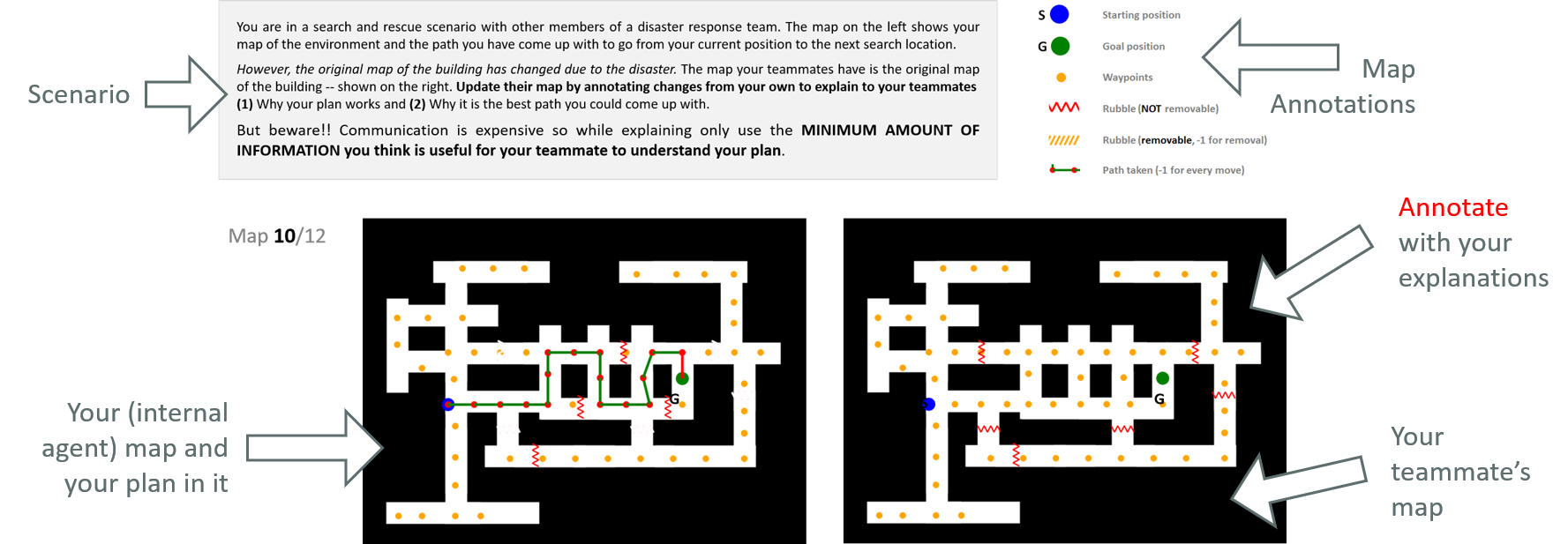}
\caption{Interface for Study-1 where participants assumed the role of the internal agent and were asked to explain their plan to a teammate with a possibly different model or map of the world.}
\label{interface1}
\end{figure*}

\section{The USAR Domain}
\label{usar}

An application where such multi-model formulations are quite useful is in typical \cite{nancy} Urban Search And Reconnaissance (USAR) tasks where a remote robot is put into disaster response operation often controlled partly or fully by an external human commander who orchestrates the entire operation.
The robot's job in such scenarios is to infiltrate areas that may be otherwise harmful to humans, and report on its surroundings as and when required / instructed by the external supervisor. 
The external usually has a map of the environment, but this map may no longer be accurate in the event of the disaster -- e.g. new paths may have opened up, or older paths may no longer be available, due to rubble from collapsed structures like walls and doors. 
The robot (internal) however may not need to inform the external of all these changes so as not to cause information overload of the commander who may be otherwise engaged in orchestrating the entire operation.
The robot is thus delegated high level tasks but is often left to compute the plans itself since it may have a better understanding of the environment.
However, the robot's actions also contribute to the overall situational awareness of the external, who may require explanations on the robots plans when necessary.
As such, such simulated USAR scenarios provide an ideal testbed for developing and 
evaluating algorithms for effective human-robot interaction. Figure~\ref{bhushi} illustrates our setup (explained in more detail in the {\bf video} (https://youtu.be/40Xol2GY7zE)). 
In the current study, we only simulate the interface to the external. This is described in details later in Section~\ref{kitty}.

In general, differences in the models of the human and the robot can manifest in any form (e.g. the robot may have lost some capability or its goals may have changed).
In the current setup, we deal with differences in the map of the environment as available to the two agents, i.e. these differences can then be compiled to differences only in the initial state of the planning problem (the human model has the original unaffected model of the world). 
This makes no difference to the underlying explanation generation algorithm \cite{explain} which treats all model changes equally.

\section{Study -- 1}
\label{arnold}

Study-1 aims to develop an understanding of how humans respond to the task of generating explanations, i.e. if left to themselves, humans preferred to generate explanations similar to the ones enumerated in Section~\ref{gloss}.
To test this, we asked participants to assume the role of the internal agent in the explanation process and explain their plans with respect to the faulty map of their teammate. 

\subsection{Experimental Setup}

Figure \ref{interface1} shows an example map and plan provided to a participant. 
On left side, the participant is shown the actual map along with the plan, starting position and the goal. 
The panel on the right shows the map that is available to the explainee. 
The maps have rubbles (both removable and non-removable) blocking access to certain paths. 
The maps may disagree as to the locations of the debris.
The participants were told that they need to convince the explainee of the correctness and optimality of the given plan by updating the latter's maps with annotations they felt were relevant in achieving that goal. We ran the study with two conditions -- 

\begin{itemize}
\item[C1.] Here the participants were asked to ensure, via their explanations, that their plan was (1) correct and (2) optimal in the updated model of their teammate; and
\item[C2.] Here, in addition to C1, they were also asked to use the minimal amount of information they felt was needed to achieve the condition in C1.
\end{itemize}

Each participant was shown how to annotate on an example (not an actual explanation) map and was then asked to explain 12 different plans using similar annotations. 
After each participant finished with the set of maps provided to them, they were provided with the following set of subjective questions -- 

{\small
\begin{verbatim}
Q1. Providing map updates were necessary to explain my plans.
Q2. Providing map updates were sufficient to explain my plans.
Q3. I found that my plans were easy to explain.
\end{verbatim}
}

The answers to these questions were measured using a five-point Likert scale.
The answers to the first two questions will help to establish whether humans 
considered map updates (or in general updates on the model differences) at all 
necessary and/or sufficient to explain a given plan.
The final question measures whether the participants found the explanation process using model differences tractable. 
It is important to note that in this setting we do not measure the efficacy of these explanations (this is the subject of Study-2 in Section~\ref{kitty}). 
Rather we are trying to find whether a human explainer would have naturally participated in the model reconciliation approach during the explanation process.

In total, we had 12 participants for condition C1 and 10 participants for condition C2 including 7 female and 18 male participants between the age range of 18-29 (data corresponding to 5 participants who misinterpreted the instructions had to be removed, 2 participants did not reveal their demographics).
Participants for the study were recruited by requesting the department secretary to send an email to the student body to ensure that they had no prior knowledge about the study or its relevance. 
Each participant was paid \$10 for taking part in the study. 

\subsection{Results}

\begin{figure}
\begin{subfigure}[b]{0.45\columnwidth}
	\includegraphics[width=\columnwidth]{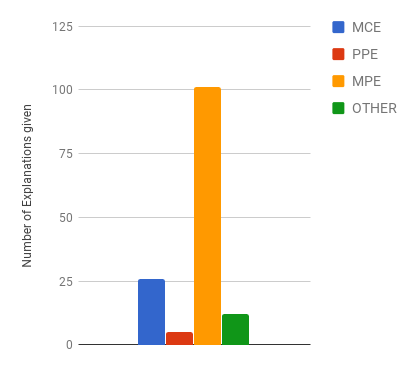}
	\caption{Study-1:C1}
	\label{free-explanations}
\end{subfigure}
\begin{subfigure}[b]{0.45\columnwidth}
	\includegraphics[width=\columnwidth]{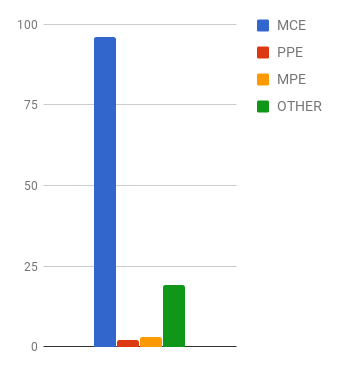}
	\caption{Study-1:C2}
	\label{min-explanations}
\end{subfigure}
\caption{Count of different types of explanations for Study-1 conditions C1 and C2.}
\label{study1}
\end{figure}

\begin{figure}
\includegraphics[width=\columnwidth]{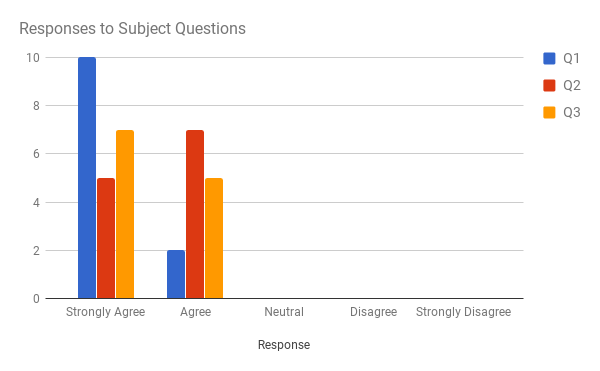}
\caption{Subjective responses of participants in Study-1:C1.}
\label{free-subj}
\end{figure}

\begin{figure}
\includegraphics[width=\columnwidth]{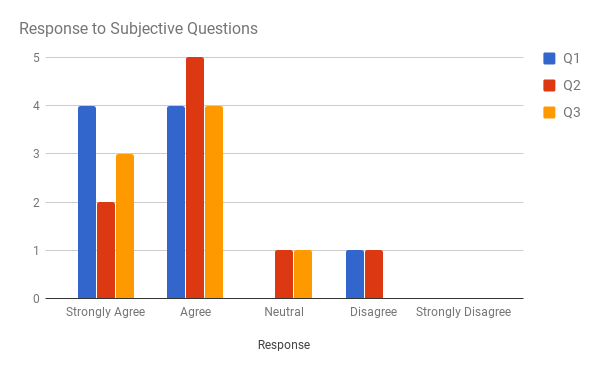}
\caption{Subjective responses of participants in Study-1:C2.}
\label{min-subj}
\end{figure}

The results of the study are presented in Figures~\ref{study1}, \ref{free-subj} and \ref{min-subj}. We summarize some of the major findings below --

\vspace{5pt}
\noindent {\bf Figure~\ref{study1} --}
The first hypothesis we tested was whether the explanations generated by the humans matched any of the explanation types discussed in Section~\ref{gloss}. 
We did this by going through all the individual explanations provided by the participants and then categorizing each explanation to one of the four types, namely MCE, PPE, MPE or Other (the "other" group contains explanations that do not correspond to any of the predefined explanation types -- more on this later in Section~\ref{study1-disc}). 
Figure~\ref{free-explanations} shows the number of explanations of each type that were provided by the participants of C1. The graph shows a clear preference for MPE. It seemed most participants felt that it was preferable to provide all the model differences.
A possible reason for this may be since the size of MPEs for the given maps were not too large (and participants did not have time constraints). 
Interestingly, in case of C2 we see a clear shift in preferences (as seen in Figure \ref{min-explanations}) where most participants ended up generating MCE style explanations. This means at least for scenarios where there are constraints on communication, the humans would prefer generating MCEs as opposed to explaining all the model differences. 

\vspace{5pt}
\noindent {\bf Figures~\ref{free-subj} and \ref{min-subj} --}
These show the results of the subjective questions for C1 and C2 respectively. Interestingly, in C1, while most people agreed on the necessity of explanations in the form of model differences, they were less confident regarding the sufficiency of such explanations. In fact, we found that many participants left additional explanations in their worksheet in the form of free text (we discuss some of these findings in Section~\ref{study1-disc}).  
In C2, we still see that more people are convinced about the necessity of these explanations than sufficiency. But we see a reduction in the confidence of the participants, which may have been caused by the additional minimization constraints.

\begin{figure*}
\includegraphics[width=\textwidth]{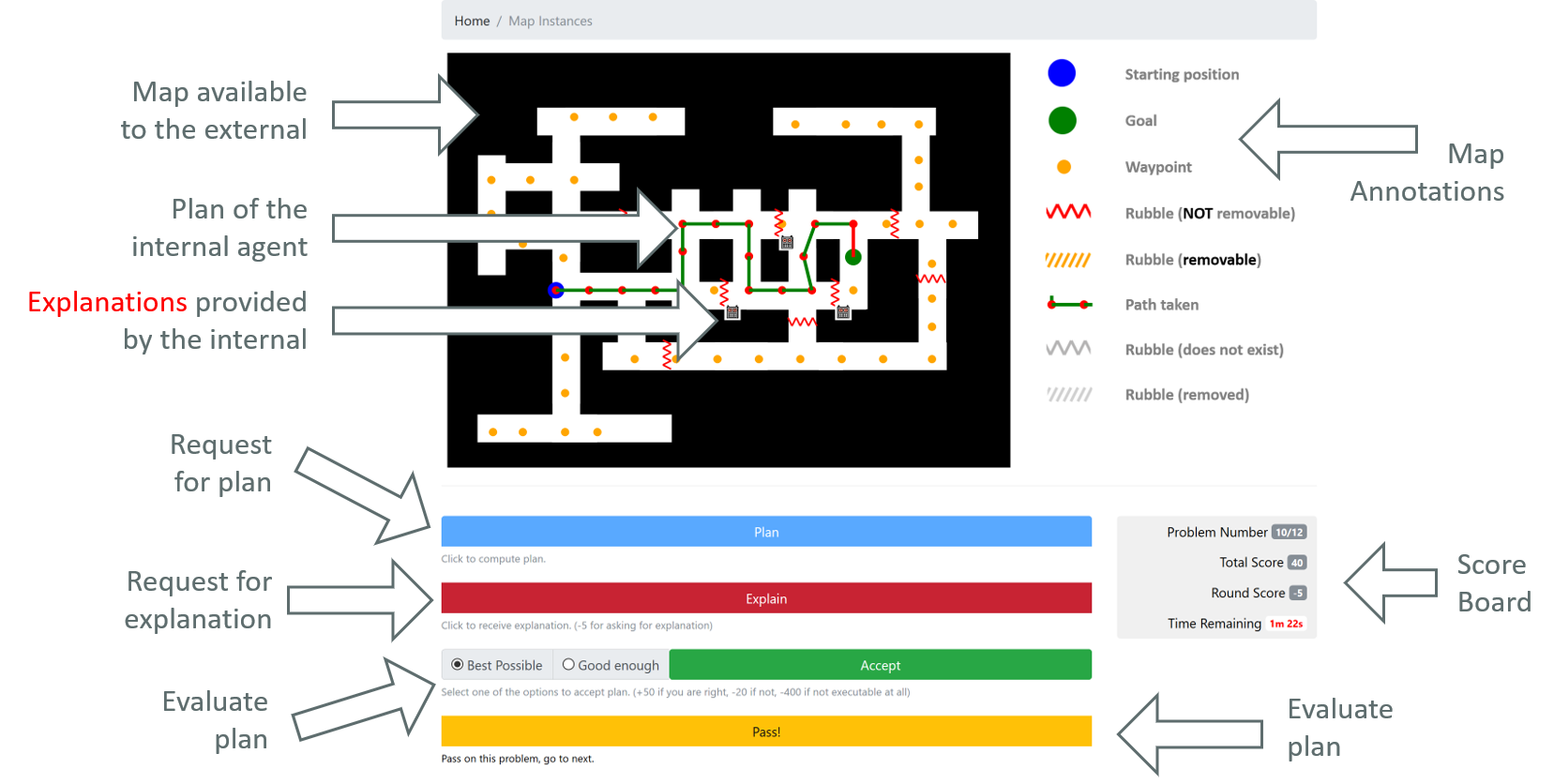}
\caption{Interface for Study-2 where participants assumed the role of the external commander and evaluated plans provided by the internal robot. They could request for plans and explanations to those plans (e.g. if not satisfied with it) and rate those plans as optimal or suboptimal based on that explanation. If still unsatisfied with the plan even after the explanation they could chose to pass and move on to the next problem.}
\label{interface2}
\end{figure*}

\section{Study -- 2}
\label{kitty}

In this study we want to understand how the different kinds of explanations outlined in Section~\ref{gloss} are perceived or evaluated when they are presented to the participants.
This study was designed to provide clues to how humans comprehend explanations when provided to them in the form of model differences. 

\subsection{Experimental Setup}

During this study, participants were incentivized to make sure that the explanation does indeed help them understand the optimality and correctness of the plans in question by formulating the interaction in the form of a game. 

Figure~\ref{interface2} shows a screenshot of the interface. 
The game displays to each participant an initial map (which they are told may differ from the robot's actual map), the starting point and the goal. 
Once the player asks for a plan, the robot responds with a plan illustrated in the form of a series of paths through the various waypoints highlighted on the map. 
The goal of the participant is to identify if the plan shown is optimal or just satisficing. 
If the player is unsure of the path, they can always ask for an explanation from the robot. 
The explanation is provided to the participant in the form of a set of model changes in the player's map. 
If the player is still unsure, they can just click on the pass button to move to the next map.

The scoring scheme for the game is as follows. 
Each player is awarded 50 points for correctly identifying the plan as either optimal or satisficing. 
Incorrectly identifying an optimal plan as just satisficing or a suboptimal plan as optimal would cost them 20 points. 
Every request for explanation would further cost them 5 points, while skipping a map does not result in any penalty. 
The participants were additionally told that selecting an inexecutable plan as either feasible or optimal would result in a penalty of 400 points. 
Even though there were no actual incorrect plans in the dataset, this information was provided to deter participants from taking chances with plans they did not understand well.

\begin{figure*}
\includegraphics[width=\textwidth]{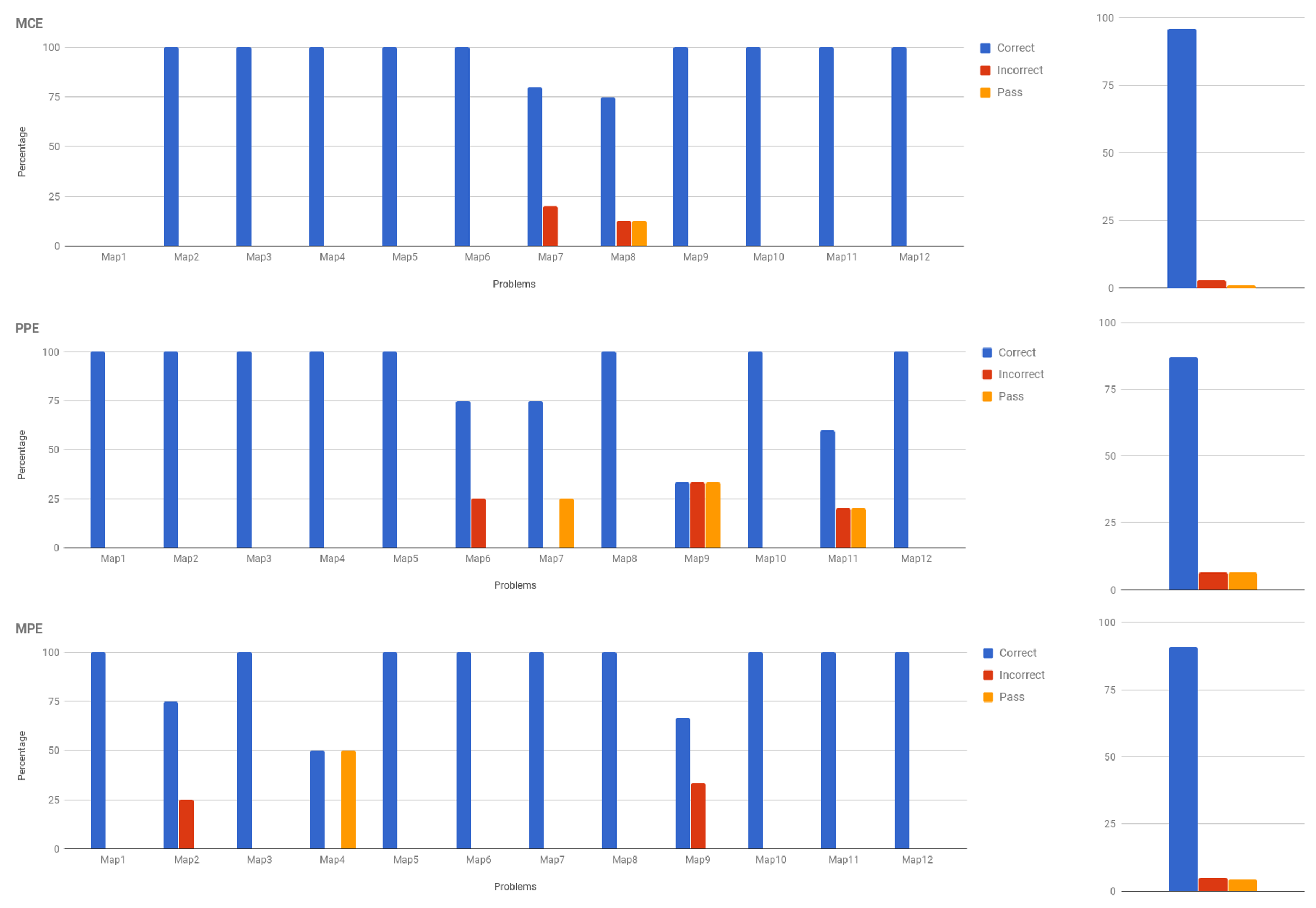}
\caption{Plots showing percentage of times different kinds of explanations (i.e. MCE / MPE / PPE) led to the correct decision on the human's part in each problem (the aggregated result is shown on the right). A ``correct decision for the human" involves them recognizing optimality of the robot's plan on being presented an MCE or an MPE, and optimality or executability (as the case may be) in case of a PPE.
}
\label{kitty-main}
\end{figure*}

Each participant was paid \$10 dollars and received additional bonuses based on the following payment scheme --

\begin{itemize}
\item[-] Scores higher than or equal to 540 were paid \$10.
\item[-] Scores higher than 540 and 440 were paid \$7.
\item[-] Scores higher than 440 and 340 were paid \$5.
\item[-] Scores higher than 340 and 240 were paid \$3.
\item[-] Scores below 240 received no bonuses.
\end{itemize}

The scoring systems for the game was designed to make sure -- 

\begin{itemize}
\item Participants should only ask for an explanation when they are unsure about the quality of the plan (due to small negative points on explanations).
\item Participants are incentivized to identify the feasibility and optimality of the given plan correctly (large reward and penalty on doing this wrongly).
\end{itemize}

Each participant was shown a total of 12 maps (same maps as in Study--1). 
For 6 of the 12 maps, the player was assigned the optimal robot plan, and when they asked for an explanation, they were randomly shown either MCE, PPE or MPE explanation with regards to the robot model. 
For the rest of the maps, participants could potentially be assigned a plan that is optimal in the human model (i.e. an explicable plan) or somewhere in between as introduced in Section~\ref{balance} (referred henceforth as the balanced plan). 
In place of the robot's optimal plan\footnote{Note that out of the 6 maps, only 3 had both balanced plans as well as explicable plans, the other 3 either had a balanced plan or the optimal human plan}. 
The participants that were assigned the optimal robot plan were still provided an MCE, PPE or MPE explanation, otherwise they were provided either the shorter explanation (for balanced plans) or an empty explanation (for the explicable plan). 
Also note that for 4 out of the 12 maps the PPE explanation cannot prove the optimality of the plan.

At the end of the study, each participant was presented with a series of subjective questions. The responses to each question were measured on a five-point Likert scale.

{\small
\begin{verbatim}
Q1. The explanations provided by the robot was helpful.
Q2. The explanations provided by the robot was easy to understand.
Q3. I was satisfied with the explanations.
Q4. I trust the robot to work on its own.
Q5. My trust in the robot increased during the study.
\end{verbatim}
}

In total, we had 27 participants for Study--2, including 4 female and 22 male participants between the age range of 19-31 (1 participant did not reveal their demographic).

\begin{figure}
\includegraphics[width=\columnwidth]{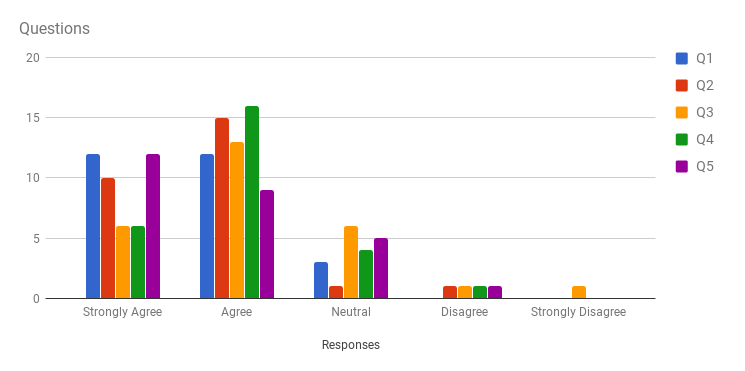}
\caption{Subjective responses of participants in Study--2.}
\label{kitty-subj}
\end{figure}

\begin{figure}[tbp!]
\includegraphics[width=\columnwidth]{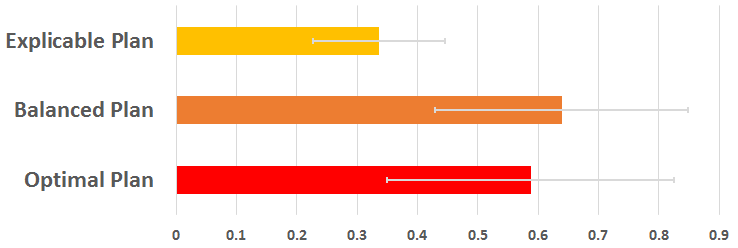}
\caption{Percentage of times explanations were sought for in Study--2 when participants presented with explicable plans vs. balanced or robot optimal plans with explanations.}
\label{kitty-percentage}
\end{figure}


\subsection{Results}

The results of the study are presented in Figures~\ref{kitty-main}, \ref{kitty-subj} and \ref{kitty-percentage}. We summarize some of the major findings below --

\vspace{5pt}
\noindent {\bf Figure~\ref{kitty-main} --}
The goal of this study is to identify if explanations in the form of model reconciliation can convince humans the optimality and correctness of the plans. 
As mentioned earlier we ran the test on 27 participants, and each participant was shown 12 unique maps and each map was assigned a random explanation type (and in some cases different plans). 
We wanted to identify whether the participants that asked for explanations were able to come up with the correct conclusions. 
This means that the subjects who asked for MCE and MPE were able to correctly identify the plans as the most optimal ones, while the people who received PPE were able to correctly classify the plan to either optimal or satisficing (i.e. for all but 5 maps PPE should be enough to prove optimality of the plans).

Figure \ref{kitty-main} shows the statistics of the selections made by participants who had requested an explanation. 
The right side of the graph shows the percentage (for every map instance) of participants who selected the correct options (marked in blue), the incorrect ones 
(marked in red) or simply passed (marked in orange), while the left side shows the average across all 12 maps. 
We notice that in general people were overwhelmingly able to identify the correct choice. 
Even in the case of PPEs, where the explanations only ensured correctness (map instances 1, 2, 3, 8 and 11) the participants were able to make the right choice. 
This shows that explanations in the form of model reconciliation are a feasible form of explanation to convey the correctness and optimality of robot plans. 
Additionally, participants can differentiate between complete and non-complete explanations based on such explanations.

\vspace{5pt}
\noindent {\bf Figure~\ref{kitty-subj} --}
These conclusions are further supported by the results from the subjective questionnaire (Figure \ref{kitty-subj}). We can see that most people seem to agree that the explanations were helpful and easy to understand. In fact, the majority of people strongly agreed that their trust of the robot increased during the study.

\vspace{5pt}
\noindent {\bf Figure~\ref{kitty-percentage} --}
We were also curious about the usefulness of explicable plans (i.e plans that are optimal in human's model) and if the humans would still ask for explanations when presented with explicable plans. 
Figure \ref{kitty-percentage} shows the percentage of times people asked for explanations when presented with various types of plans.
These results are from 382 problem instances that required explanations, and 25 and 40 instances that contained balanced and explicable plans respectively. 
From the perspective of the human, the balanced plan and the robot optimal plan do not make any difference since both of them appear suboptimal. 
This is evident from the fact that the click-through rate for explanations in these two conditions are similar.  
However, the rate of explanations is considerably less in case of explicable plans as desired.
This matches the intuition behind the notion of plan explicability as a viable means (in addition to explanations) of dealing with model divergence in human-in-the-loop operation of robots.

It is interesting to see that in Figure~\ref{kitty-percentage} about a third of the time participants still asked for explanations even when the plan was explicable, and thus optimal in their map. This is an artifact of the risk-averse behavior incentivized by the gamification of the explanation process and indicative of the cognitive burden on the humans who are not (cost) optimal planners. Thus, going forward, the objective function should incorporate the cost or difficulty of analyzing the plans and explanations from the point of view of the human in addition to the current costs of explicability and explanations (as shown in Table~\ref{tab1}) modeled from the perspective of the robot model (refer to \cite{exp-fss} for more details).

\vspace{5pt}
\noindent {\bf Table~\ref{tab1}} shows the statistics of the explanations / plans from  124 problem instances that required {\em minimal} explanations as per \cite{explain}, and 25 and 40 instances that contained balanced and explicable plans respectively, as before.
As desired, the robot gains in length of explanations but loses out in cost of plans produced as it progresses along the spectrum of optimal to explicable plans. Thus, while Table~\ref{tab1} demonstrates the cost of explanation versus explicability trade-off from the robot's point of view, Figure~\ref{kitty-percentage} shows how this trade-off is perceived from the human's perspective.

\begin{figure}[tbp!]
\includegraphics[width=\columnwidth]{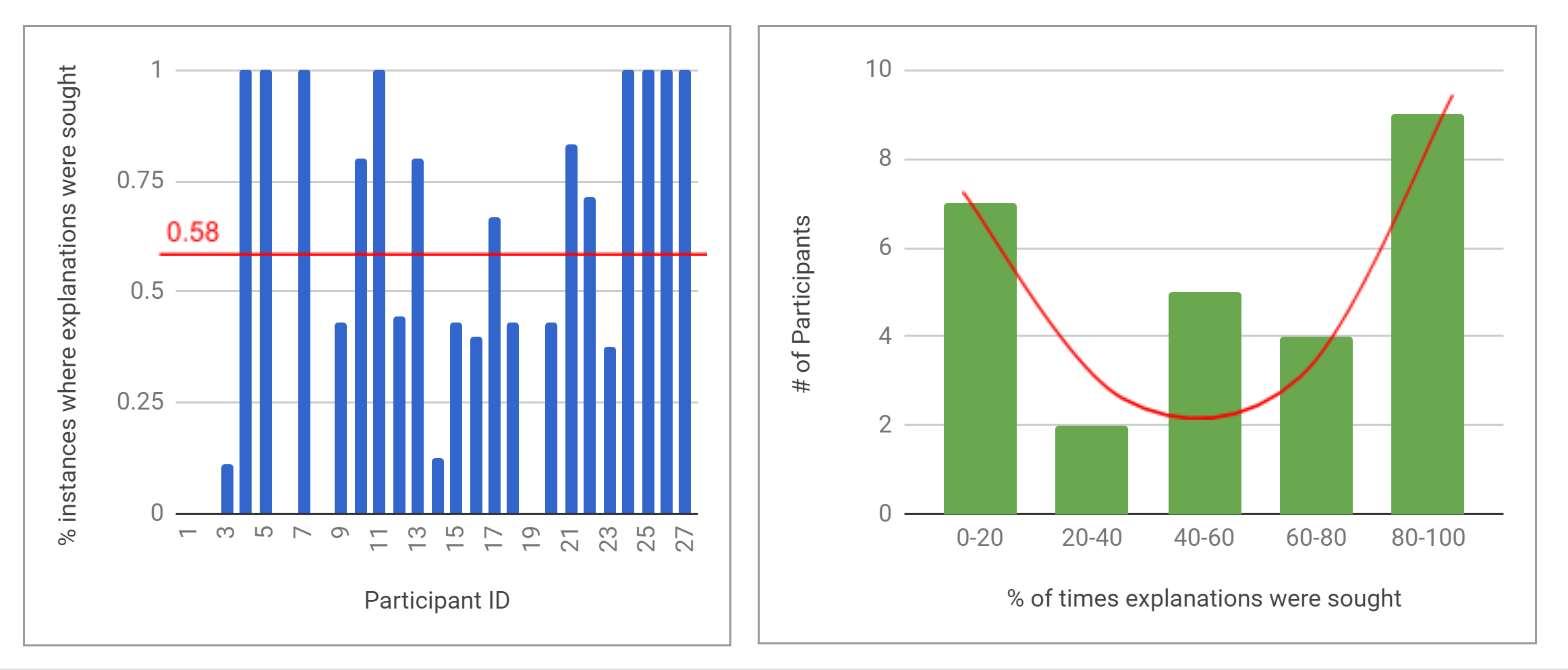}
\caption{Click-through rates for explanations in Study--2 revealing either risk-taking behavior or risk-averse behaviors from the participants.}
\label{kitty-huh}
\end{figure}

\vspace{5pt}
\noindent {\bf Figure~\ref{kitty-huh}} shows how the participants responded to inexplicable plans, in terms of their click-through rate on the explanation request button. Such information can be used to model the $\alpha$ parameter to situate the explicability versus explanation trade-off according to preferences of individual users. 
It is interesting to see that the distribution of participants (right inset) seem to be bimodal indicating that there are people who are particularly skewed towards risk-averse behavior and others who are not, rather than a normal distribution of response to the explanation-explicability trade-off. This further motivates the need for learning $\alpha$ interactively with the particular human in the loop.

\begin{table}[tbp!]
\centering 
\begin{tabular}{|c|c||c|c||c|c|}
\hline
\multicolumn{2}{c}{Optimal Plan} & \multicolumn{2}{c}{Balanced Plan} & \multicolumn{2}{c}{Explicable Plan} \\
\hline
$|\mathcal{E}|$ & $C(\pi, \mathcal{M}^R)$ & $|\mathcal{E}|$ & $C(\pi, \mathcal{M}^R)$ & $|\mathcal{E}|$ & $C(\pi, \mathcal{M}^R)$ \\
\hline
2.5 & 5.5 & 1 & 8.5 & - & 16\\
\hline
\end{tabular}
\vspace{10pt}
\caption{Statistics of explicability versus explanation trade-off with respect to explanation length and plan cost.}
\label{tab1}
\end{table}

\section{Discussions}
\label{study1-disc}

As we mentioned before, there were some instances where the participants from Study 1 generated explanations that are outside the scope of any of the explanation types discussed in Section~\ref{gloss}. These are marked as ``Other'' in Figure~\ref{study1}.
In the following we note three of these cases that we found interesting --

\subsection{Post-hoc explanations}

Notice that parts of an MCE that actually contribute to the executability of a given plan may not be explained in post-hoc situations where the robot is explaining plans that have {\em already been done} as opposed to plans that are being proposed for execution.
The rationale behind this is that if the human sees an action, that would not have succeeded in his model, actually end up succeeding (e.g. the robot had managed to go through a corridor that was blocked by rubble) then he can rationalize that event by updating his own model (e.g. there must not have been a rubble there). 
This seems to be a viable approach to further reduce size (c.f. selective property of explanations in \cite{miller}) of explanations in a post-hoc setting, and is out of scope of explanations developed in \cite{explain}.

\subsection{Identification of Explicit Foils}

Identification of explicit foils \cite{explain} can help reduce the size of explanations as well. In the explanations introduced in Section~\ref{gloss} the foil was implicit -- i.e. why this plan {\em as opposed to all other plans}. However, when the implicit foil can be estimated (e.g. top-$K$ plans expected by the human) then the explanations can only include information on why the plan in question is better than those other options (which are either not executable or costlier). Some participants provided explanations contrasting some of these foils in terms of (and in addition to just) the model differences.

\subsection{Cost-based reasoning}

Finally, a kind of explanation that was attempted by some participants involved a cost analysis of the current plan with respect to foils (in addition to model differences, as mentioned above). Such explanations have been studied extensively in previous planning literature \cite{danmaga,dave} and seems to be still relevant for plan explanations on top of the model reconciliation process.


\section{Conclusion}
The paper details the results of studies aimed to evaluate the effectiveness of plan explanations in the form of model reconciliation. 
Through this study, we aimed to validate whether explanations in the form of model reconciliation (in its various forms) suffice to explain the optimality and correctness of plans to the human in the loop. 
We also studied cases where participants were asked to generate explanations in the form of model changes, to see if explanations generated by the humans align with any of the explanations identified in existing literature. 
The results of the study seem to suggest that humans do indeed understand explanations of this form and believe that such explanations are necessary to explain plans.
In future work, we would like to investigate how explanations must be adapted for scenarios where the robot is expected to interact with the humans over a long period of time and how such interactions affect the dynamics trust and teamwork in human-robot collaborations.

\bibliographystyle{ACM-Reference-Format}
\bibliography{bib} 

\end{document}